\documentclass[journal,twoside,web]{IEEEtran}
\usepackage{xcolor}
\usepackage{tmi}
\usepackage{cite}
\usepackage{amsmath,amssymb,amsfonts}
\usepackage[ruled,norelsize,vlined,linesnumbered]{algorithm2e}
\makeatletter
\newcommand{\removelatexerror}{\let\@latex@error\@gobble}
\makeatother
\usepackage{multirow}
\usepackage{booktabs}
\usepackage{graphicx}
\usepackage{textcomp}
\usepackage{amssymb}
\usepackage{hyperref}

\def\BibTeX{{\rm B\kern-.05em{\sc i\kern-.025em b}\kern-.08em
    T\kern-.1667em\lower.7ex\hbox{E}\kern-.125emX}}
\begin{document}
\title{Enriched text-guided variational multimodal knowledge distillation network for automated diagnosis of plaque in 3D carotid artery MRI}
\author{
Bo Cao, Fan Yu, Mengmeng Feng, SenHao Zhang, Xin Meng, Yue Zhang, Zhen Qian, and Jie Lu
\thanks{This work was supported in part by  National Natural Science Foundation of China grant No. 82130058 and No. 81974261. (corresponding authors: Jie Lu)}
\thanks{Bo Cao, Fan Yu, Mengmeng Feng, SenHao Zhang, Xin Meng, Yue Zhang, and Jie Lu are with the department of Radiology and Nuclear Medicine, Xuanwu Hospital, Capital Medical University, and also with the Beijing Key Laboratory of Magnetic Resonance Imaging and Brain Informatics, Beijing 100053, China (e-mail: init$\_$cb@163.com; imaginglu@hotmail.com).}
\thanks{Zhen Qian is with Beijing United Imaging Research Institute of Intelligent Imaging, Beijing, China (e-mail: zhenqian@gmail.com)}
}

\maketitle

\begin{abstract}
Multimodal learning has attracted much attention in recent years due to its ability to effectively utilize data features from a variety of different modalities. Diagnosing the vulnerability of atherosclerotic plaques directly from carotid 3D MRI images is relatively challenging for both radiologists and conventional 3D vision networks. In clinical practice, radiologists assess patient conditions using a multimodal approach that incorporates various imaging modalities and domain-specific expertise, paving the way for the creation of multimodal diagnostic networks. In this paper, we have developed an effective strategy to leverage radiologists' domain knowledge to automate the diagnosis of carotid plaque vulnerability through Variation inference and Multimodal knowledge Distillation (VMD). This method excels in harnessing cross-modality prior knowledge from limited image annotations and radiology reports within training data, thereby enhancing the diagnostic network's accuracy for unannotated 3D MRI images. We conducted in-depth experiments on the dataset collected in-house and verified the effectiveness of the VMD strategy we proposed.
\end{abstract}

\begin{IEEEkeywords}
Multimodal learning, Knowledge distillation, Atherosclerosis plaque, 3D MRI
\end{IEEEkeywords}

\section{Introduction}
\label{sec:introduction}
\IEEEPARstart{P}ATIENTS with carotid stenosis often have associated carotid atherosclerotic plaques, among which vulnerable plaques carry the risk of rupture, potentially leading to ischemic stroke events with high disability and mortality rates. \cite{cheng2019contemporary,collaborators2021global,halliday2004group,rothwell2003analysis,howard2021risk}.
High-resolution magnetic resonance imaging sequences can effectively display the degree of carotid stenosis and detailed plaque composition, such as thin-cap fibroatheroma (TCFA), inflammatory cell infiltration, lipid-rich necrotic core (LRC), calcification (CA), and intraplaque hemorrhage (IPH)\cite{2019PlaqueVulnerable,cai2002classification,brunner2021associations,bos2021atherosclerotic}. 
Precise analysis of plaque risk level and vulnerability typically requires years of diagnostic experience by radiologists to identify dangerous components such as LRC and IPH within mixed compositions, guiding the choice of subsequent treatment methods\cite{bonati2022management,chen2021domain}.

\begin{figure}
    \centering
    \includegraphics[width=1\linewidth]{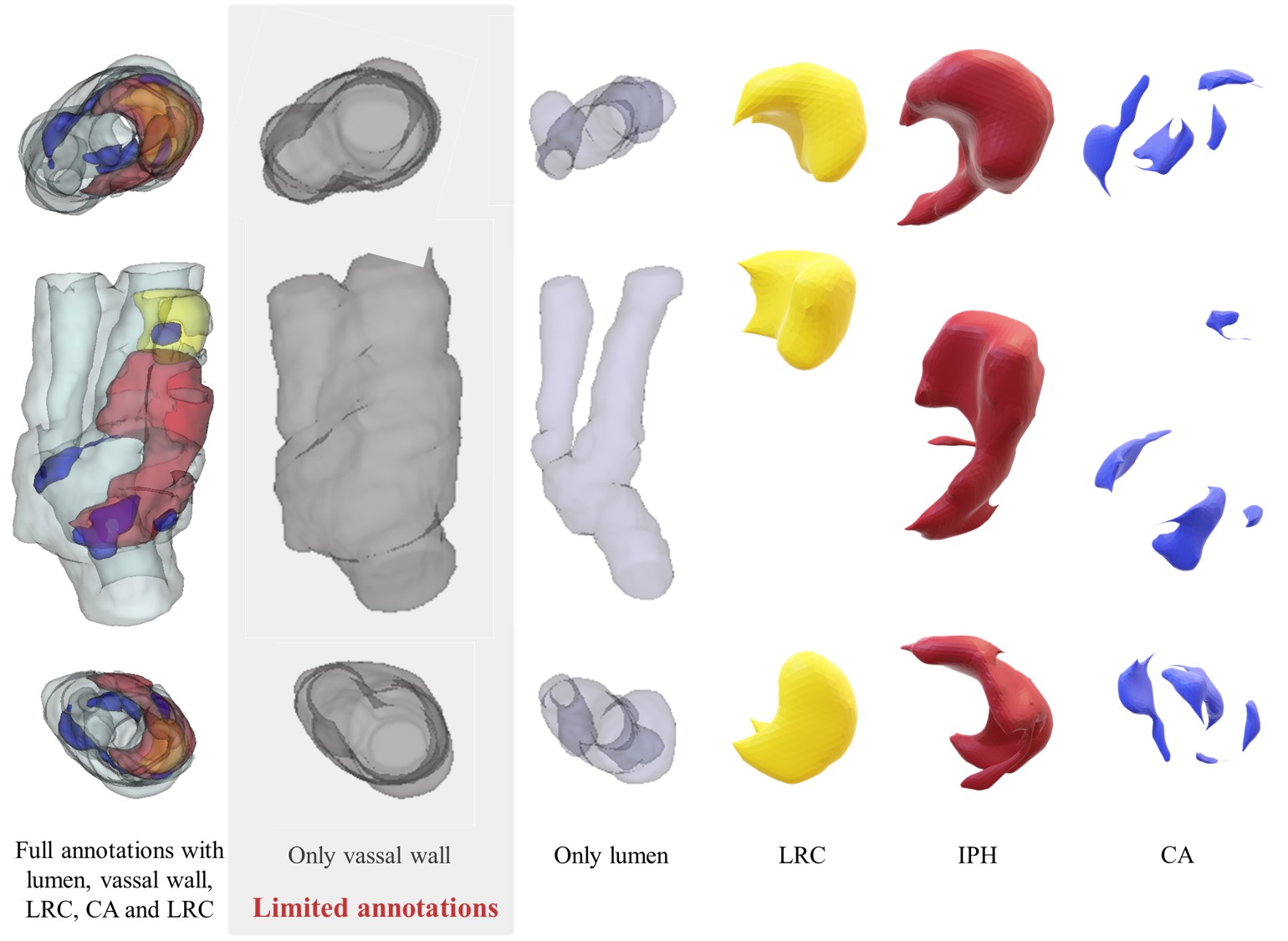}
    \caption{Example of the difference between full annotation and limited annotation. Full annotation images include multiple complex annotations of the carotid vessel wall, lumen, lipid-rich necrotic core (LRC), calcification (CA), and intraplaque hemorrhage (IPH), requiring detailed pixel-level annotations. Limited annotation only involves annotating the vessel region at the carotid bifurcation,without the need to involve the lumen and other components.}
    \label{fig:annotation}
\end{figure}

\begin{figure}
    \centering
    \includegraphics[width=1\linewidth]{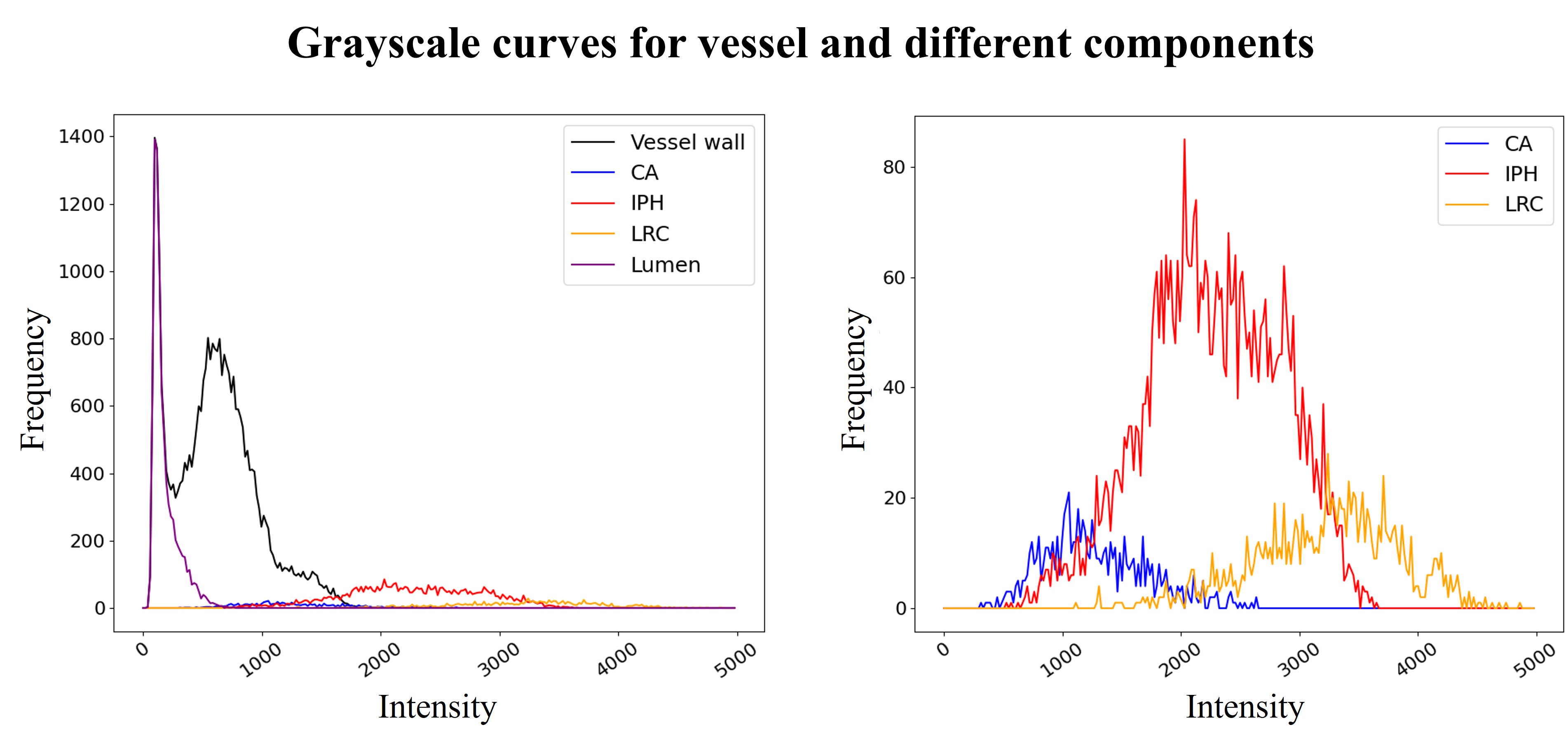}
    \caption{Example of grayscale overlaps for different components and limited annotation with vessel wall. The left image shows the grayscale distribution of the vessel wall, lumen, and various complex components within the plaque in the MRI image. The grayscale coverage is obtained by calculating the ratio of the lumen and component parts to the entire vessel area. The right image displays the grayscale distribution histograms of CA, LRC, and IPH. Despite the differences, the boundaries of the grayscale values remain quite ambiguous, posing challenges for annotation and segmentation.}
    \label{fig:grayscale}
\end{figure}

U-Net-like visual networks\cite{U-net,zhou2019unet++,hatamizadeh2022unetr} have been employed for tasks such as vessel wall segmentation and component detection to address the challenges of carotid vessel analysis\cite{WANG2024yufan,chen2021domain,wu2019deep}. 
However, these strongly supervised segmentation strategies often entail significant labor costs due to intensive medical image annotation, especially on 3D medical images, and the subjective nature of these annotations can introduce errors that complicate clinical applications\cite{WANG2024yufan}. 
One solution is to adopt limited annotations without detailed plaque component information to reduce labor intensity, as shown in Fig. \ref{fig:annotation}, but this may limit the model’s ability to directly assess plaque vulnerability from the segmentation results. However, due to the specificity of plaques and their components, detailed annotations require labeling of fine components with a grayscale coverage exceeding 30\%. Moreover, the grayscale differences between vulnerable components such as LRC and IPH, which have different signal expressions in the images, and relatively stable components like CA, are quite ambiguous, as shown in Fig. \ref{fig:grayscale}.

Considering that the additional anatomical information provided by limited annotations, such as the location of carotid plaques, is crucial in the diagnostic process, the method of knowledge distillation(KD)\cite{KD,CRD} inspires us to treat these limited annotations as an additional modality. 
Through the distillation process, the learnable features related to anatomical structures can be transferred to the diagnostic target model. 
Therefore, it is crucial to maximize the informational richness derived from limited annotations during the training phase.

It is noteworthy that multimodal information produced during the diagnostic process, such as radiology reports, contains rich domain knowledge, providing a natural environment for constructing multimodal learning\cite{tiu2022expert,CLIP,VKD}, yet it is often overlooked in the automatic analysis of carotid plaques. 
However, the modal differences between language and vision may introduce noise that is detrimental to the construction of diagnostic models\cite{BioLinkBERT,clinicalbert}. 
Therefore, it becomes highly appealing to enable neural networks to learn expert information from cross-modal information that is beneficial for precise classification in unannotated carotid 3D MRI images while eliminating noise interference.

In this paper, we developed a novel three-level distillation strategy based on \textbf{V}ariation inference and \textbf{M}ultimodal knowledge \textbf{D}istillation(\textbf{VMD}) to transfer comprehensive understanding of radiology reports and limited annotations to student network for plaque vulnerability diagnosis by a student-teacher-expert structure, which enabled us to accomplish the complex task at a lower cost. 

\begin{figure}
    \centerline{\includegraphics[width=1\linewidth]{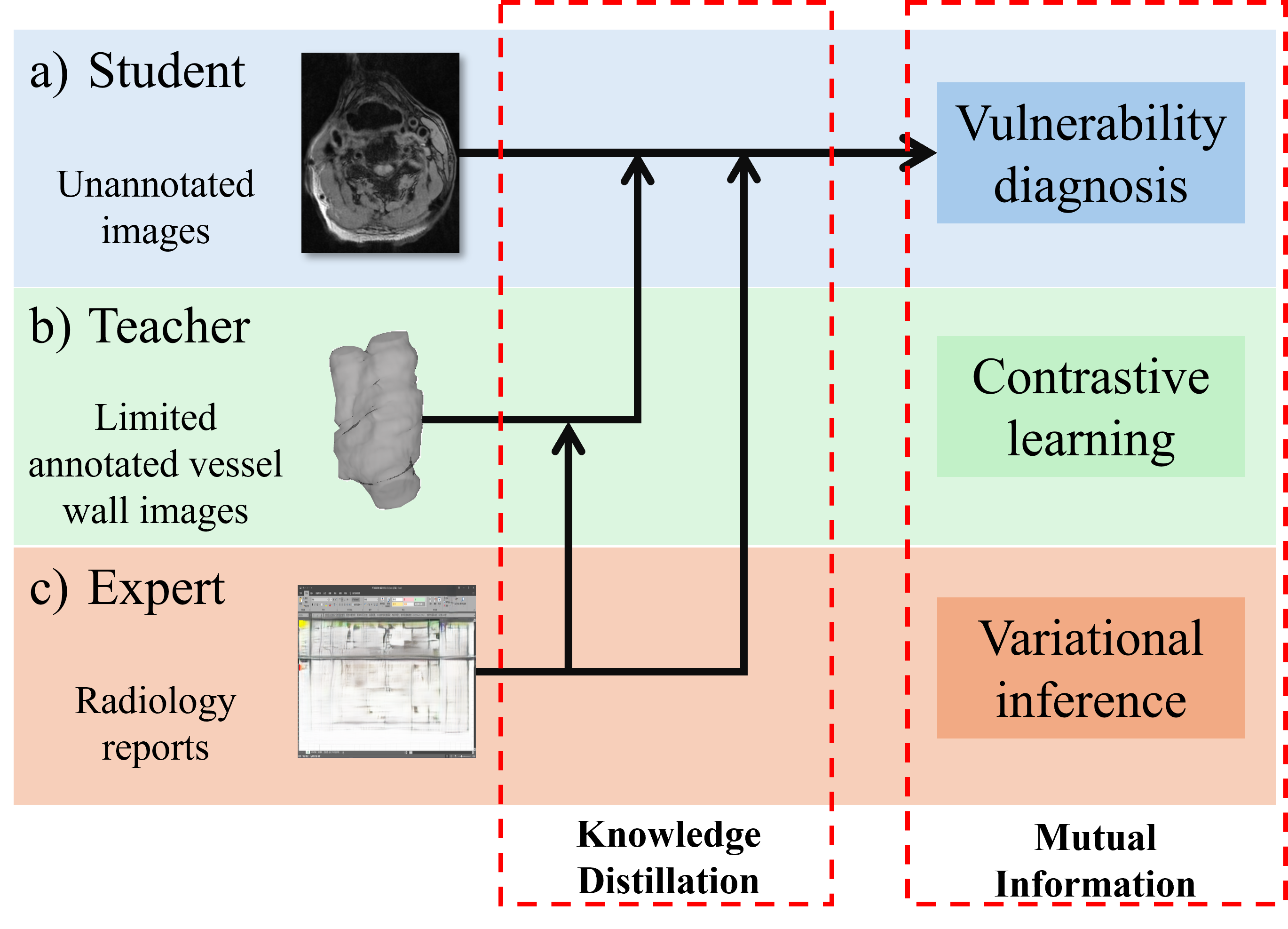}}
\caption{\textbf{How limited annotations and reports work in VMD.}
The diagnosis of plaque vulnerability is directly performed using unannotated 3D MRI by the student network (a). 
Limited annotations and radiology reports, which act as teacher (b) and expert (c) respectively, influence the student by maximizing contrastive learning-based MI and variational inference. 
This process significantly enhances the diagnostic capabilities of the student.
    \label{figure:simple_model}}
\end{figure}

Contrastive learning-based mutual information (MI) optimization excels in effective cross-modal feature transfer and alignment, similar to KD\cite{KD,NCEloss,CRD,CCL,AMID}. 
Limited annotated data, focusing on the vessel area, provide noise-reduced anatomical information for diagnosing plaque vulnerability and could be considered an additional modality for multimodal learning. 
Exploiting the similarity between annotated and unannotated data, we speculate that samples of the same category, whether they are vulnerable or stable plaques, from these two modalities should exhibit a higher degree of similarity\cite{CCL,AMID}. 
Therefore, we have developed a contrastive learning-based KD network aimed at improving the diagnostic accuracy for unannotated images. 
This is achieved by maximizing MI between the student model, which lacks annotations, and the teacher model, which has access to limited annotations. 
This approach is designed to leverage the inherent similarities within the data to enhance the learning process, enabling the student model to better identify and diagnose plaque vulnerability without the need for extensive annotated data, as shown in Fig. \ref{figure:simple_model}(b).

The expertise comes from radiology reports that can be integrated into the aforementioned networks, offering expert-level guidance to both the teacher and student models\cite{BioLinkBERT,clinicalbert}. Therefore, we have designed an additional expert network to improve the previous distillation strategy for both the teacher and student networks. The KD method based on variational inference does not require additional costs at the inference stage \cite{VKD,VAE}, making it suitable for direct diagnosis from unannotated images. This approach, which minimizes a specific Kullback-Leibler Divergence (KLD) to estimate the evidence lower bound (ELBO) of the optimization objective\cite{VAE}, facilitates the transfer of knowledge across different modalities and can be considered as another way of maximizing MI, as shown in Fig. \ref{figure:simple_model}(c).

\section{Related works}
\subsection{Student-Teacher distillation structure}
The concept of KD was first proposed in \cite{KD} and is considered highly effective in model compression and feature transfer. The student-teacher distillation structure used therein has been applied in many complex tasks. 
VID \cite{VID} maximizes the MI between the student and the teacher network by minimizing the entropy of the teacher to effectively transfer knowledge to the student.
CRD \cite{CRD} introduces contrastive learning into the student-teacher structure. 
CCL \cite{CCL} separates image and audio in video classification tasks and optimizes the performance through feature mixing contrast learning. This idea is further explained in AMID \cite{AMID} from the perspective of MI among different modalities. 
However, our method, i.e., VMD, utilizes manual annotations to acquire anatomical knowledge from the unannotated images and feed it back to the student by maximizing inter-class MI.

\subsection{Image-Text Medical Knowledge Transfer}
Pre-trained biomedical semantic understanding models have demonstrated commendable proficiency in comprehending medical texts in recent years \cite{Mc-BERT,clinicalbert,BioLinkBERT}. 
Some studies \cite{VKD,tiu2022expert,textGuided} guided the neural network in diagnosing diseases in 2D image modality by extracting features closely associated with the diagnosis from text by a pre-trained language model.
These methods employ either knowledge distillation \cite{sohn2015learning,KD} or contrastive learning \cite{CLIP} to align features in order to address the problem. 
By integrating contrastive learning with conditional variational inference, we expanded the strategy of VKD \cite{VKD} into a student-teacher-expert structure to provide a tighter constraint to transfer the knowledge from the expert to teacher networks because of the substantial disparity between 3D and 2D images.

\subsection{3D Carotid MRI automatic analysis}
Rapid and precise intelligent analysis is highly appealing for the clinical diagnosis of carotid stenosis and atherosclerotic plaque MRI images. Some studies have been designed to address this issue. Wu \textit{et al.}\cite{wu2019deep} proposed an end-to-end DeepMAD network that performed excellently in the task of carotid vessel wall segmentation in black-blood MRI sequences, achieving high-precision segmentation of the vessel lumen and outer wall. Zhu \textit{et al.}\cite{zhu2021cascaded} designed a carotid segmentation model for multi-parameter MRI based on a 3D residual U-net and a multi-level joint feature optimization strategy. Chen \textit{et al.}\cite{chen2021domain} proposed a carotid vessel segmentation strategy based on HR-MRI, which can effectively segment the lumen and vessel wall at the carotid bifurcation from 2D MRI images, achieving classification of normal vessels, early lesions, and advanced lesions, but did not involve the division of specific plaque components. Wang \textit{et al.}\cite{WANG2024yufan} designed an end-to-end multi-parameter black-blood MRI carotid vessel segmentation and plaque component recognition network with good performance in component recognition, which had its application range somewhat limited by the need for multi-parameter MRI images. Li \textit{et al.}\cite{shendinggangTMI} combined multimodal data input from CTA and MR with SAM adaptation for medical images, designing a method to extend sparse annotations and segment the carotid vessel wall and lumen. The aforementioned analytical methods are largely based on labor-intensive manual annotations. The fatigue caused by prolonged detailed annotations may introduce inconsistencies in annotation quality, and they overlook the potential role of radiology report information that contains rich guiding prior knowledge and is easily accessible.

\begin{figure*}
\centering
\includegraphics[width=473pt]{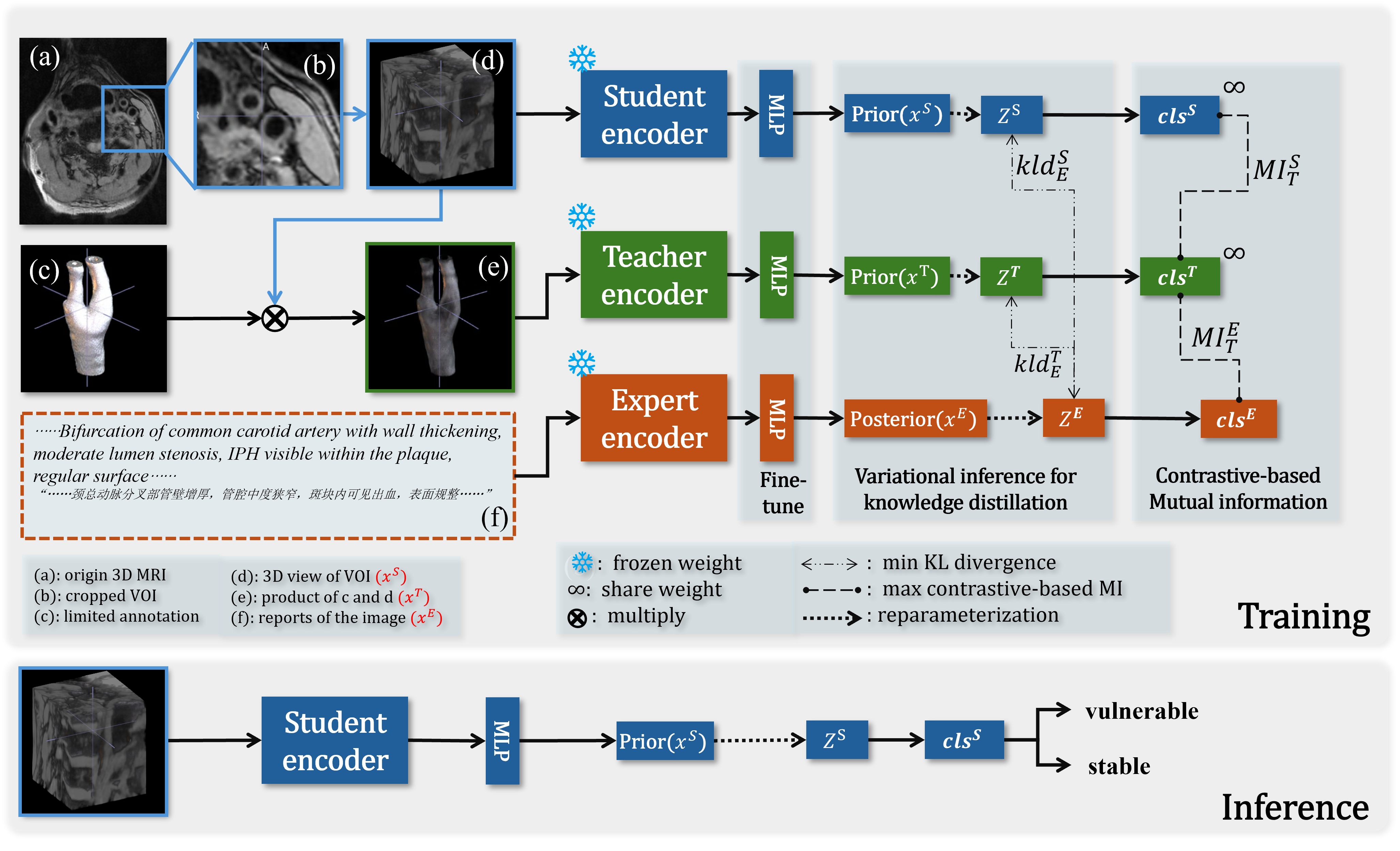}
\caption{\textbf{The overview of  VMD.} (a) is the transverse section of the original 3D MRI image. The carotid artery region (b,d) for the student network is obtained through cropping. The input (e) for the teacher network is the product of (d) and the limited annotations (c). The radiology report (f) corresponding to the image is for the expert network. When training VMD, we maximize the MI and minimize the KL divergence between the latent spaces $z$ of different networks.
During the inference phase, there is no guidance for the student to diagnose the plaque vulnerability.}
\label{figure:full_model}
\end{figure*}

\section{Methods}
The overall architecture of VMD is illustrated in Fig. \ref{figure:full_model}. 
We employed pre-trained models with frozen weights as feature encoders for the Expert network $E$ (dark orange), teacher network $T$ (green), and the weaker-performing student network $S$ (blue), and fine-tuned them using different fully connected layers. 
The primary objective was to reduce the cost of training the network in the early stages of learning 3D images and report modalities.
This strategy has been repeatedly proven effective in medical imaging tasks.Then we apply amortization techniques and reparameterization  \cite{VAE} to get the prior$\left(x^{T}\right)$ and prior$\left(x^{S}\right)$ for both $T$ and $S$. We use a two-layer MLP whose latent feature size is 512 to play the role of $z$. Using techniques akin to those for image input, we compute the posterior$\left(x^{E}\right)$.

In order to efficiently transfer cross-modal prior knowledge, we need to maximize the MI between different student-teacher pairs \cite{CRD}.
During the optimization process, we also minimize the cross-entropy $H_{cls}$ of all networks as much as possible to make the prediction approach the true value. The global optimization objective $\mathcal{O}$ is shown in (\ref{Eq: global o}), $\alpha_1$,$\alpha_2$,$\alpha_3$,$\alpha_4$ are weights for the corresponding terms.
\begin{equation}\label{Eq: global o}
    argmax(\mathcal{O}) = \alpha_1\mathcal{I}\left(S,T\right)+\alpha_2\mathcal{I}\left(S,E\right)+\alpha_3\mathcal{I}\left(T,E\right)-\alpha_4H_{cls}
\end{equation}
where the $\mathcal{I}$ means MI. We will sequentially solve for each term of the equation in the following sections.

\subsection{Cross-modal knowledge transfer from reports: $\mathcal{I}\left(S,E\right)$}
Conditional variational inference \cite{sohn2015learning} is highly suitable for structured prediction tasks, and VKD \cite{VKD} has validated its effectiveness in understanding medical image analysis tasks. 
We construct a variational probability model and formulate the problem of predicting the target $y$ from the unannotated image $x^{S}$ as solving the posterior probability $p\left( y \middle| x_{I}^{S} \right)$. In order to obtain the following conditional likelihood formula, we introduce the latent variable $z^{S}$:

\begin{equation}
\begin{aligned}
    {\log{p\left( y \middle| x_{I}^{S} \right)}} = {\log{\int{p\left( y \middle| {x_{I}^{S},z_{I}^{S}} \right)p\left( z_{I}^{S} \middle| x_{I}^{S} \right)dz_{I}^{S}~}}}
\end{aligned}
\end{equation}

the $p\left( z_{I}^{S} \middle| x_{I}^{S} \right)$is the conditional prior for $x^{S}$, which is easy to obtain by the neural network, as shown in Fig. \ref{figure:full_model} denoted as prior$\left(x^{S}\right)$.
The prediction problem can be translated into maximizing this integral w.r.t. $z$. 
Thus, we need to find the maximized posterior probability  $p\left( z^{S}\middle|x^{S},y\right)$, which is relatively difficult to solve. Inspired by \cite{VKD}, we introduce the representation $z^{E}$ from the expert network and use the variational posterior distribution  $q\left(z^{E}\middle|x^{E}\right)$ to approximate $p\left( z^{S}\middle|x^{S},y\right)$. 
Given that the KLD is well-behaved in the distance metrics \cite{VAE}, we can obtain $kld^{S}_{E}$ in Fig. \ref{figure:full_model} by applying Bayesian formulae given the weight $\theta$ of the networks, and translating the original problem into minimizing the KLD between $q\left(z^{E}\middle|z^{E}\right)$ and $p\left( z^{S}\middle| x^{S} \right)$:

\begin{equation}
\begin{aligned}
 kld^{S}_{E} 
 &=D_{KL}\left\lbrack q_{\theta}\left(z^{E}\middle|x^{E}\right) \middle| \left| {p\left( {z^{S}} \middle| {x^{S},y} \right)} \right\rbrack \right.
    \\&=\int{q_{\theta}\left(z^{E}\middle|x^{E}\right)
        \log{
            \frac{q_{\theta}\left(z^{E}\middle|x^{E}\right)
            }{p\left( {z^{S}} \middle| {x^{S},y} \right)
            }
        }
    }d\theta
    \\&=\int{q_{\theta}\left(z^{E}\middle|x^{E}\right)\log{q_{\theta}\left(z^{E}\middle|x^{E}\right)}}d\theta
    \\&+\int{q_{\theta}\left(z^{E}\middle|x^{E}\right)\log{p\left(y\middle|x^{S}\right)}}d\theta
    \\&-\int{q_{\theta}\left(z^{E}\middle|x^{E}\right)\log{p{\left(z^{S}\middle|x^{S},y\right)}}}d\theta
    \\&=\log{p\left(y\middle|x^{S}\right)}+\int{q_{\theta}\left(z^{E}\middle|x^{E}\right)\log{p\left(z^S\middle|x^S,y\right)}}d\theta
    \\&+\int{q_{\theta}\left(z^{E}\middle|x^{E}\right)\log\frac{p\left(y\middle|x^S\right)}{q_{\theta}\left(z^{E}\middle|x^{E}\right)}}d\theta
    \\&={\log{p\left(y\middle|x^{S}\right)}} - \mathbb{E}_{q_{\theta}\left(z^{E}\middle|x^{E}\right)}\left\lbrack {\log{p\left( y \middle| {x^{S},z^{S}} \right)}} \right\rbrack
     \\&+ D_{KL}\left\lbrack q_{\theta}\left(z^{E}\middle|x^{E}\right) \middle| \middle| p\left( z^{S}\middle| x^{S} \right) \right\rbrack \\
\end{aligned}
\end{equation}
Since $D_{KL}>=0$, we can obtain an upper bound ${\log{p\left(y\middle|x^{S}\right)}}$, and assuming that it is a constant, we can approximate it by optimizing the networks and the latter is the optimization objective, as shown in (\ref{Eq:elbo 0}). 
\begin{equation}\label{Eq:elbo 0}
\begin{aligned}
    {\log{p\left(y\middle|x^{S}\right)}} &>= \mathbb{E}_{q_{\theta}\left(z^{E}\middle|x^{E}\right)}\left\lbrack {\log{p\left( y \middle| {x^{S},z^{S}} \right)}} \right\rbrack
     \\&- D_{KL}\left\lbrack q_{\theta}\left(z^{E}\middle|x^{E}\right) \middle| \middle| p\left( z^{S}\middle| x^{S} \right) \right\rbrack \\
\end{aligned}
\end{equation}
from where we can maximize the $\mathcal{I}\left(S,E\right)$ by maximizing the $ELBO_{SE}$ as follows:
\begin{equation}\label{Eq:elbo 1}
\begin{aligned}
    \mathcal{I}\left(S,E\right) &= ELBO_{SE} = \mathbb{E}_{q\left(z^{E}\middle|x^{E}\right)}\left\lbrack {\log{p\left( y \middle| {x^{S},z^{S}} \right)}} \right\rbrack \\&- D_{KL}\left\lbrack q\left(z^{E}\middle|x^{E}\right) \middle| \middle| p\left( z^{S}\middle| x^{S} \right) \right\rbrack \\
\end{aligned}
\end{equation}

\subsection{Maximizing mutual information based on comparative learning: $\mathcal{I}\left(S,T\right)$}
For the homologous annotated $x^{T}$ and unannotated $x^{S}$, suppose a pair of predictions $\hat{x}^{T}_{j}$ and $\hat{x}^{S}_{i}$ derived from $T$ and $S$, we aim to minimize the distance between predictions with the same class label and maximize the gap between predictions with different classes for all $N$ samples. To maximize the MI between $S$ and $T$, we consider applying a similarity-based model $ M\left( {  \hat{x}^{T}_{j}, \hat{x}^{S}_{i}} \right)$ inspired by \cite{AMID} to quantify the difference between $\hat{x}^{T}_{j}$and $\hat{x}^{S}_{i}$, shown in (\ref{Eq:mi}).
\begin{equation}\label{Eq:mi}
    M\left( {  \hat{x}^{T}_{j}, \hat{x}^{S}_{i}} \right) = \frac{\exp\left( {\phi\frac{ \hat{x}^{T}_{j}, \hat{x}^{S}_{i}}{\tau}} \right)}{\sum\limits_{k = 1}^{N}{\exp\left( {\phi\frac{{\left(\hat{x}^{T}_{j}, \hat{x}^{S}_{k}\right)}}{\tau}} \right)}}
\end{equation}
where $\phi$ is a cosine similarity scoring function, and temperature $\tau$ is set empirically to $0.5$. For $B$ samples of the same category out of all $N$ samples, we choose to optimize the $\mathcal{I}\left(S,T\right)$ in a form that looks like infoNCE loss \cite{NCEloss} but has additional supervision guided by class label, as shown in (\ref{Eq:MIst}),

\begin{equation}\label{Eq:MIst}
\begin{aligned}
    \mathcal{I}\left(S,T\right) &=\frac{1}{N}\sum_{j=1}^{N}\sum_{i=1}^{B}\log{ M\left( {\hat{x}^{T}_{j},\hat{x}^{S}_{i}} \right)} \\&+ \frac{1}{N}\sum_{j=1}^{N}\sum_{k=1}^{N-B}\log\left(1-M\left( { \hat{x}^{T}_{j},\hat{x}^{S}_{k}} \right)\right)
\end{aligned}
\end{equation}

\begin{figure}
    \centering
    \includegraphics[width=1\linewidth]{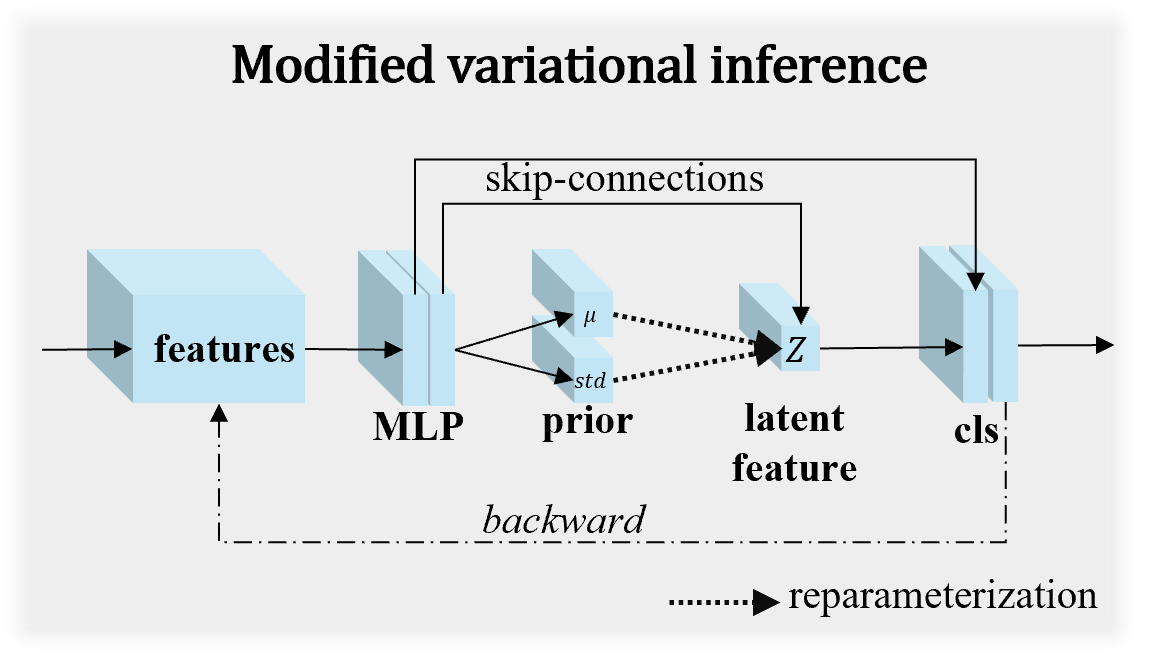}
    \caption{Example of the modified variational inference module. The Student, Teacher, and Expert share similar inference structure modules, fine-tuning feature parameters from different pre-trained networks through a two-layer MLP, calculating the mean and standard deviation respectively, and then ensuring gradient backpropagation through the reparameterization trick.}
    \label{fig:Variational inference}
\end{figure}

\begin{figure}
    \centering
    \includegraphics[width=1\linewidth]{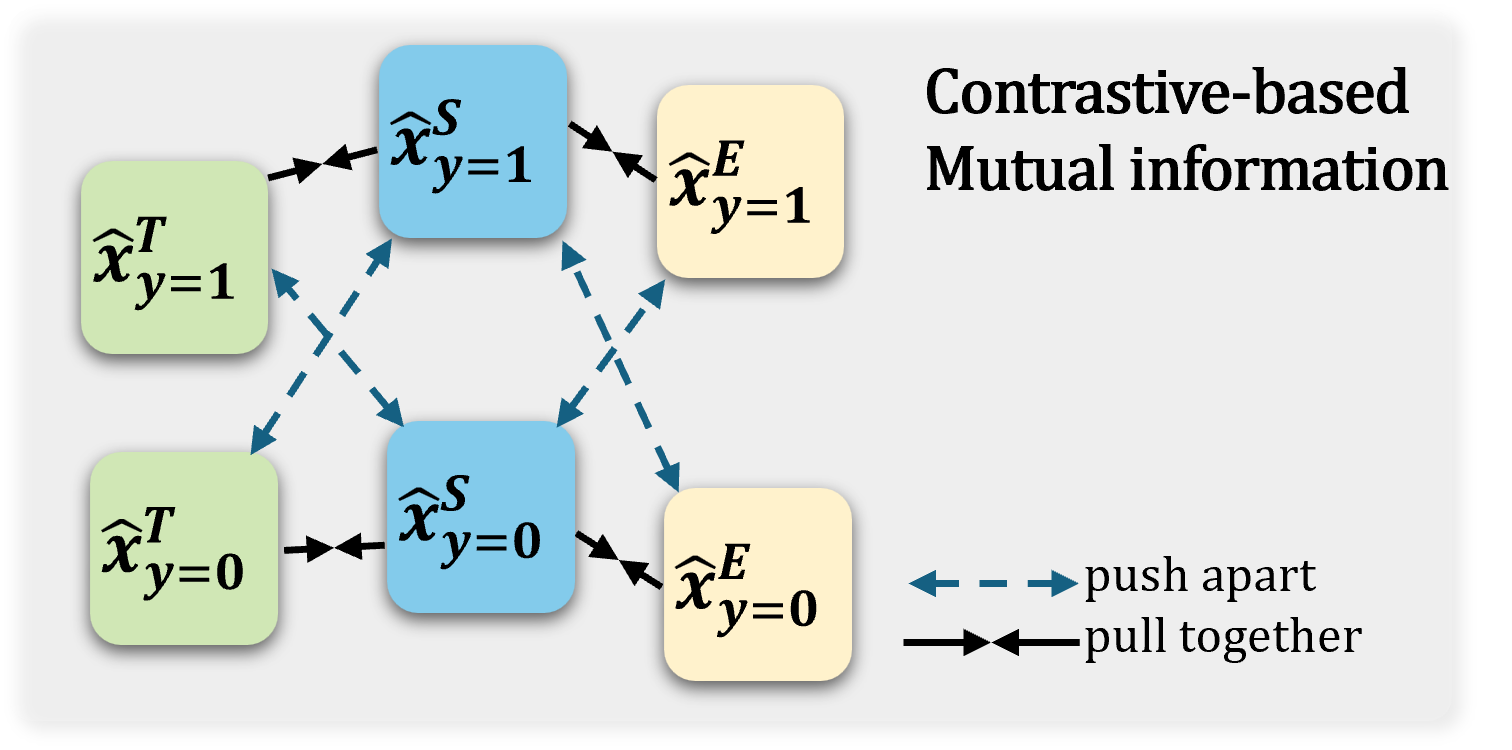}
    \caption{Example of the contrastive-based inter-class mutual information optimization strategy. The Student’s predicted class-level output for the input sample is further constrained by the Teacher, which can be viewed as aligning the underfitted distribution curve with the better-fitted feature curve through similarity optimization. The Expert’s constraint on the Teacher follows the same principle.}
    \label{fig:Contrastive-based MI}
\end{figure}
\subsection{The further constraint on the teacher: $\mathcal{I}\left(T,E\right)$}
The further constraint on the teacher stems from both variational inference and contrastive learning. Therefore, we first compute the KLD between the latent spaces $z$ based on (\ref{Eq:elbo 1}) and then compute the similarity between the predictions based on (\ref{Eq:MIst}). It is noteworthy that $T$ and $S$ share the same parameters in the classification head. 

Equation (\ref{Eq: teacher}) shows the  summary of $\mathcal{I}\left(T,E\right)$.

\begin{equation}\label{Eq: teacher}
\begin{aligned}
   \mathcal{I}\left(T,E\right) 
   &= \mathbb{E}_{q\left(z^{E}\middle|x^{E}\right)}\left\lbrack {\log{p\left( y \middle| {x^{T},z^{T}} \right)}} \right\rbrack \\&- D_{KL}\left\lbrack q\left(z^{E}\middle|x^{E}\right) \middle| \middle| p\left( z^{E}\middle| x^{T} \right) \right\rbrack \\&+\frac{1}{N}\sum_{j=1}^{N} \sum_{i=1}^{B}\log{ M\left( { \hat{x}^{T}_{j},\hat{x}^{E}_{i}} \right)} \\&+\frac{1}{N}\sum_{j=1}^{N} \sum_{k=1}^{N-B}\log\left(1-M\left( { \hat{x}^{T}_{j},\hat{x}^{E}_{k}} \right)\right)
\end{aligned}
\end{equation}

 At last, we use the minimized cross-entropy loss function $\mathcal{L}_{ce}$ to further obtain more accurate classifications for all 3 parts of VMD, as shown in (\ref{Eq:cls}), $\lambda_1$,$\lambda_2$,$\lambda_3$ are different weights for these terms. 
 \begin{equation}\label{Eq:cls}
     H_{cls} = \lambda{_1}\mathcal{L}_{ce}\left(\hat{x}^{T}_{i},y_{i}\right)+\lambda{_2}\mathcal{L}_{ce}\left(\hat{x}^{S}_{i},y_{i}\right)+\lambda{_3}\mathcal{L}_{ce}\left(\hat{x}^{E}_{i},y_{i}\right)
 \end{equation}

Through our reasoning, modified variational inference and contrastive-based on similarity optimization have been well integrated from the perspective of MI optimization. 

Substituting the above reasoning results into (\ref{Eq: global o}) yields the final optimization objective of VMD.

\begin{table*}
\caption{Detail descriptions of train, validation, and test set.\label{table:detail dataset}}%
\begin{tabular*}{\textwidth}{@{\extracolsep\fill}cccccccccccccccc@{\extracolsep\fill}}
\toprule
\multirow{2}{*}{\textbf{Dataset}} & \multirow{2}{*}{\textbf{Number}} & \multirow{2}{*}{\textbf{Age$^{1}$}} & \multicolumn{2}{c}{\textbf{Gender}} & \multicolumn{3}{c}{\textbf{Location$^{2}$}}     & \multicolumn{3}{c}{\textbf{Stenosis degree$^{3}$}}        & \multicolumn{5}{c}{\textbf{Components$^{4}$}}                               \\ \cmidrule(l){4-16} 
                                  &                                  &                               & \textbf{male}   & \textbf{female}   & \textbf{CB} & \textbf{ICA} & \textbf{CCA} & \textbf{mild} & \textbf{moderate} & \textbf{severe} & \textbf{LRC} & \textbf{CA} & \textbf{IPH} & \textbf{UC} & \textbf{TB} \\ \cmidrule(r){1-3}
\textbf{train}                    & 338                              & 65.76±6.96                    & 294             & 44                & 305         & 29           & 4            & 131           & 152               & 55              & 166          & 178         & 63           & 36          & 0           \\
\textbf{val}                      & 82                               & 65.11±7.02                    & 70              & 12                & 77          & 5            & 0            & 31            & 45                & 6               & 42           & 44          & 21           & 22          & 2           \\
\textbf{test}                     & 82                               & 62.8±7.54                     & 78              & 4                 & 77          & 3            & 2            & 28            & 38                & 16              & 45           & 40          & 16           & 11          & 3           \\ \bottomrule

\end{tabular*}

\begin{itemize} 
    \item[1] Age in years, displayed as average age with standard deviation.
    \item [2] The location of plaques, CB is carotid bifurcation, ICA is internal carotid artery, CCA is common carotid artery.
    \item[${3}$] Mild stenosis: $<$30\%; Moderate stenosis: 30\%-69\%; Severe stenosis: 70\%-99\%;
    \item[${4}$] The components of plaques, LRC is lipid-rich necrotic core, CA is calcification, IPH is intraplaque hemorrhage, UC is ulceration, and TB is thrombus.
\end{itemize}
    
\end{table*}

\begin{figure}[!t]
 \removelatexerror
  \begin{algorithm}[H]
   \caption{Training of VMD}\label{alg:train}
   \KwIn{$x^S_i$, $x^E_i$, $l_i$. $i =  1, ... ,N$ , label $l_n$ which contains vessel wall generated by radiologists.}
  \KwOut{predictions $\hat{x}^{S}_{i}, \hat{x}^{T}_{i}, \hat{x}^{E}_{i}$ for three submodels.}
   Get $x^T_i$ by the product of $x^S_i$ and $l_i$(see Fig. \ref{figure:full_model}(e))\;
   \For( \emph{$i := 1$ to $N$}){if not converged}
   {
      Compute prior($x^{S}$), prior($x^T$), and posterior($x^E$) by student encoder, teacher encoder, and expert encoder, respectively\;
      Draw latent features $z^S$, $z^T$, $z^E$ using amortization techniques and reparameterization\cite{VAE}\;
      Minimise $ kld^{S}_{E} $ and $ kld^{T}_{E} $ to transfer the knowledge from radiologist reports to the vision encoders. (see (\ref{Eq:elbo 1}) and (\ref{Eq: teacher}))\;
      Predict the class label $\hat{x}^S_i$, $\hat{x}^T_i$, and $\hat{x}^E_i$ of the input images and reports from the classification heads. (see (\ref{Eq:cls}))\;
      Maximize the mutual information $M\left({\hat{x}^{T}_{j}, \hat{x}^{S}_{i}}\right)$ and $M\left({\hat{x}^{T}_{j},\hat{x}^{E}_{i}}\right)$ between the expert-teacher and teacher-student pairs to further constrain the distribution of features by leveraging anatomical structures(see (\ref{Eq:MIst} and (\ref{Eq: teacher}))\;
       Update all weights of VMD by gradient descent\;
   }
   
  \end{algorithm}
\end{figure}

\begin{figure}[!t]
 \removelatexerror
  \begin{algorithm}[H]
   \caption{Inference of VMD's student}\label{alg:test}
   \KwIn{original images $x^S_i$. $i =  1, ... ,N$.}
  \KwOut{predictions $\hat{x}^{S}_{i}$.}
   \For( \emph{$i := 1$ to $N$}){}
   {
      Compute prior($x^{S}$) by student encoder\;
      Draw latent features $z^S$ by amortization techniques and reparameterization\cite{VAE}\;
      Predict the class label $\hat{x}^S_i$\;
      
   }
   
  \end{algorithm}
\end{figure}

\section{Experiment}
\subsection{Dataset}
Patients diagnosed with carotid stenosis according to the 2022 guidelines of the European Society for Vascular Surgery\cite{aburahma2022society}, who were hospitalized in the Department of Neurosurgery at Capital Medical University Xuanwu Hospital from September 2021 to September 2023, were retrospectively included. 
The experiment incorporated 502 bilateral carotid artery 3D-T1-FSE images from 303 patients, which included a total of 350 vulnerable images and 152 stable images.
Inclusion criteria were: (1) age 45-85 years; 
(2) presence of plaque in unilateral or bilateral carotid artery determined by ultrasonography (wall thickening $>$1.5 mm)\cite{chen2021domain}
(3) no contraindications to MR or contrast agent injection. 
Exclusion criteria were: (1) history of carotid endarterectomy, carotid stenting, or neck radiotherapy; 
(2) history of tumor or chemotherapy; 
(3) recent acute or chronic infection, autoimmune disease, or ongoing anti-inflammatory treatment; 
(4) poor image quality. 
This study adhered to the Declaration of Helsinki and was approved by the Ethics Committee of Xuanwu Hospital of Capital Medical University, exempting subjects from informed consent, with the approval number: [2022] 023.

The selected high-quality images are categorized into average quality and good quality based on the presence of issues such as the lumen and the external contour of the blood vessel being locally blurred, with few artifacts that do not affect the diagnosis. There are 153 images of average quality and 349 images of good quality.

The imaging was performed using a 3.0T PET/MR (United Imaging uPMR790) with an 8-channel carotid-specific surface coil. 
Scanning was centered on the carotid bifurcation. Sequence parameters were: TR = 800 ms, TE = 14.52 ms, FOV = 180 mm × 180 mm, resolution = 0.60 mm × 0.60 mm × 0.60 mm, number of layers = 390. 

Two experienced radiologists($>$10 years of experience in neuroradiology) collaboratively classified the images of each vessel as either “vulnerable” or “stable”, in which plaques containing LRC, IPH, thrombus, and ulceration are marked as vulnerable, as indicated in Table. \ref{table:detail dataset}.
These 486 images were all cropped to a volume of (128, 128, 60) based on the scanning center before the start of training. And the training and test sets are separated using a ratio of 5:1, and 5-fold cross-validation was employed for model training and validation,  ensuring that the independent test set was not visible during the training phase.

\subsection{Implementation}
 We first employ MedicalNet \cite{chen2019med3d}, pre-trained on 23 CT and MRI public datasets, as the visual encoder backbone network and use a two-layer MLP to fine-tune it. For the expert network $E$, we apply Mc-BERT \cite{Mc-BERT}, a Chinese medical semantic understanding BERT-base network pre-trained in a considerably large Chinese biomedical text corpus \cite{chineseBLUE}, as the Expert encoder and use one single transformer layer to fine-tune it.
All experiments were performed in two NVIDIA A40 GPUs, using the Adam optimizer for 400 epochs. The learning rate is initialized to 5e-4 and weight decay is 1e-4. Five random seeds were used to calculate the average and standard deviation of ROC and other evaluation metrics. The $\beta$ of Fbeta score is 0.5.
\subsection{Evaluations}
We compared the diagnostic performance of plaque vulnerability between VMD and other methods by constructing confusion matrices and comparing the AUCs. When comparing with junior doctors($<$2 years experience in neuroradiology), we used the professional statistical analysis software SPSS 25.0 for analysis, conducting T-tests on the diagnostic accuracy and time consumption between VMD and the doctors, with a P-value less than 0.05 considered statistically significant.
\section{Results}
\begin{figure*}
    \centering
    \includegraphics[width=1\linewidth]{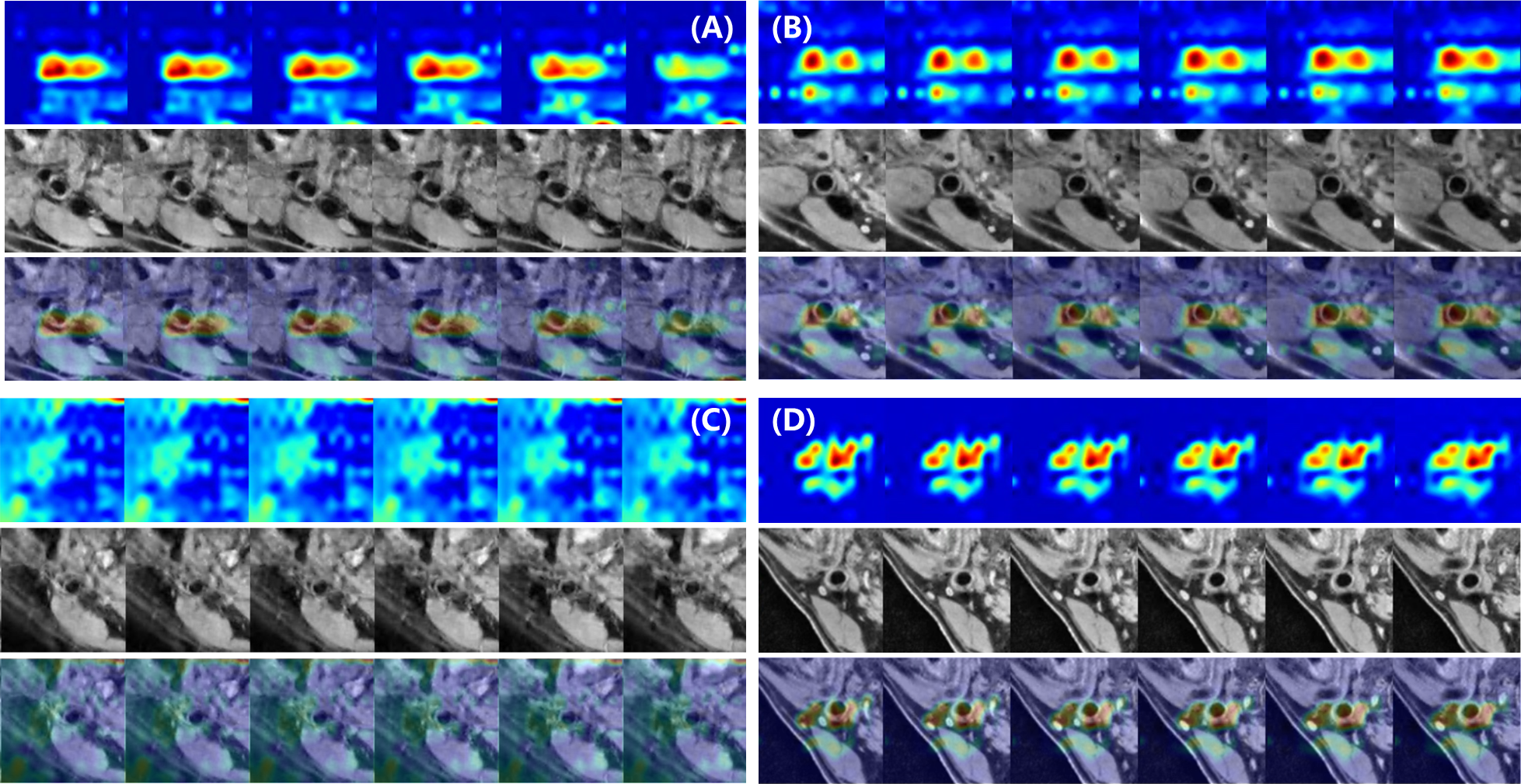}
    \caption{Utilize the Grad-CAM\cite{grad_cam_2020} method to visualize the features of the final convolutional layer of the student network. For 4 distinct test samples, we selected 6 consecutive 2D images from the middle layers to form 6 columns for display. The first row shows the heatmaps of the visualized feature distributions, the second row shows the original images, and the third row shows the overlay of both. The visualized feature distributions of images are aligned with expectations.}
    \label{fig:img_cam}
\end{figure*}

We first compare VMD with the baseline method, three state-of-the-art knowledge distillation methods, and a network fine-tuning method, namely VID\cite{VID}, VKD\cite{VKD}, CCL\cite{CCL}, and VPT\cite{VPT}, as shown in Table. \ref{table:compare}. We employed the fully fine-tuned 3DResNet-50 network as our classification baseline, a model derived from MedicalNet. Additionally, we used the grad-CAM method\cite{grad_cam_2020} to visualize the feature distribution of the terminal 3D convolutional layer in the student network of VMD during the inference diagnosis process, as shown in Fig. \ref{fig:img_cam}. Secondly, we compare our method with the diagnostic results of junior doctors, demonstrating the effectiveness of VMD in terms of diagnostic efficiency and accuracy. Finally, we conducted ablation studies to validate the significance of the teacher-expert module in VMD.
\begin{table*}
\centering
\caption{Compare classification performance with SOTA methods, \textbf{bold} represnets the best.}
\label{table:compare}
    \begin{tabular}{cccccc}
    \toprule
    \textbf{Setting}     & \textbf{ROC}           & \textbf{Accuracy}     & \textbf{Precision}     & \textbf{Recall}        & \textbf{Fbeta$^1$}         \\ 
    \midrule
    \textbf{Baseline\cite{chen2019med3d}} & 0.6567±0.0018          & 0.6588±0.0007         & 0.7802±0.0055          & 0.7213±0.0063          & 0.7487±0.003           \\
    \textbf{VPT\cite{VPT}}         & 0.6141±0.0244          & 0.6195±0.0055         & 0.7044±0.0074          & {0.7966±0.0077}   & 0.7476±0.0009          \\
    \textbf{VID\cite{VID}}         & 0.6594±0.018           & {{0.7171±0.0055}}   & 0.7959±0.0016          & \textbf{0.8069±0.0077} & \textbf{0.8014±0.0046} \\
    \textbf{VKD\cite{VKD}}         & {{0.6818±0.0162}}    & 0.678±0.0185          & 0.8065±0.0203& 0.7172±0.0094          & 0.7592±0.0123          \\
    \textbf{CCL\cite{CCL}}         & 0.659±0.0005           & 0.6361±0.0044         & 0.778±0.0045           & 0.6779±0.0123          & 0.7238±0.0035          \\
    \midrule
    \textbf{VMD}        & \textbf{0.7136±0.013} & \textbf{0.7244±0.0204} & \textbf{0.8197±0.0192} & 0.7828±0.0094          & {0.8008±0.0135}    \\
    \bottomrule
    \end{tabular}
    \begin{itemize}
        \item[$^{1}$] The $\beta$ of Fbeta score is 0.5, same below.
    \end{itemize}
\end{table*}
\subsection{Compare with SOTA methods} VPT \cite{VPT} facilitates network fine-tuning by introducing few learnable parameters into the original 2D images, but the advantage is not evident in 3D images. 
VID \cite{VID} minimizes the entropy of the teacher network to transfer knowledge from the teacher to the student. 
VKD \cite{VKD} transfers abundant knowledge from the EHRs to the X-ray diagnosis network by introducing variational knowledge distillation.
CCL \cite{CCL} introduces a contrastive learning-based MI optimization method in the cross-modal knowledge distillation problem, realizing feature transfer from audio and images to video. 
For the sake of fairness in comparative experiments, we used the fine-tuned baseline method 3DResNet-50 \cite{chen2019med3d} as the image encoder for different methods to compare. 
Table. \ref{table:compare} shows the comparison results of the classification performance of VMD and the other SOTA methods in unannotated carotid artery 3D MRI. 
We achieved more than 1\% performance improvement in the first three metrics (ROC: 4.66\%, ACC: 1.01\%, PRC: 1.63\%) and one close to SOTA, and we obtained performance gains of 8.67\%, 10.05\%, 5.06\%, 8.52\%, and 6.96\% in five metrics respectively compared to the baseline. 
ROC enhancement is critical for demonstrating the diagnostic power of the model in clinical practice, thus we think the performance of VMD for automatic diagnosis of plaque vulnerability has achieved a meaningful advancement compared to previous studies. 
\subsection{Diagnosis feature visualization}
In the prediction phase, the feature distribution of the last bottleneck layer in the image encoder of the VMD student network was visualized using the Grad-CAM\cite{grad_cam_2020}, as shown in Fig. \ref{fig:img_cam}. We selected 4 representative images and displayed the middle 6 layers of each 3D image. The first row in each of the four subplots is the heatmap, with red representing areas of high network attention and blue representing areas of low attention; the second row is the original image; the third row is the overlay of the above two images. It can be seen that the convolutional network of the student network focuses on the carotid plaque MRI images, concentrating on the carotid vessels and their surroundings, which aligns with our expectations.

\subsection{Compare with junior radiologist}
In terms of deep learning-based automated methods, we deployed our VMD network on an Nvidia A40 graphics card and obtained the trained student network as the inference model for comparative experiments. On the other hand, we included two junior radiologists for experimental comparison. We selected 54 cases from the independent test set for the experiment, including 36 3D MRI labeled as vulnerable and 18 images labeled as stable, with the distribution ratio of the two types of images being the same as the training set. The two junior radiologists were asked to diagnose 27 randomly selected images from the 54 cases. By recording the diagnostic efficiency and accuracy of both VMD and the radiologists, we demonstrated the effectiveness of our method. Statistical methods were used for significance analysis.

\begin{table*}  
    \centering
    \caption{ablation study for teacher w/o expert (\ref{Eq: teacher}) to diagnose annotated images.}
    \label{table:ablation2}
    \begin{tabular}{@{}cccccccc@{}}
\toprule
\textbf{Limited Annotations} & \textbf{Report} & \textbf{Setting}               & \textbf{ROC}           & \textbf{Accuracy}      & \textbf{Precision}     & \textbf{Recall}        & \textbf{Fbeta}         \\ \midrule
\checkmark & × & \multicolumn{1}{l}{w/o expert} & 0.7372±0.0048          & 0.6732±0.0134          & 0.8151±0.0101          & 0.6966±0.0378          & 0.7505±0.018           \\
\checkmark & \checkmark & teacher                        & \textbf{0.7923±0.0316} & \textbf{0.7829±0.0181} & \textbf{0.8222±0.0487} & \textbf{0.7945±0.0229} & \textbf{0.8079±0.0344} \\ \bottomrule
\end{tabular}
\end{table*}

\begin{table*}[t]%
    \centering
    \caption{ablation study for student w/o teacher (\ref{Eq:MIst}),(\ref{Eq: teacher}) and expert (\ref{Eq:elbo 1}),(\ref{Eq: teacher}) to diagnose unannotated images.}
    \label{table:ablation3}
        \begin{tabular}{@{}ccccccccc@{}}
    \toprule
    \textbf{(\ref{Eq:elbo 1})} & \textbf{(\ref{Eq:MIst})} & \textbf{(\ref{Eq: teacher})} & \textbf{Setting} & \textbf{ROC}           & \textbf{Accuracy}     & \textbf{Precision}     & \textbf{Recall}        & \textbf{Fbeta}         \\ \midrule
    ×             & ×             & ×             & w/o both         & 0.6645±0.0193          & 0.6537±0.0109         & 0.7859±0.0181          & 0.7034±0.0463          & 0.7412±0.0197          \\
    \checkmark             & ×             & ×             & w/o teacher      & 0.6818±0.0162          & 0.678±0.0185          & 0.8065±0.0203          & 0.7172±0.0094          & 0.7592±0.0123          \\
    ×             & \checkmark             & ×             & w/o expert       & 0.6815±0.028           & 0.6512±0.0164         & 0.76±0.0175            & 0.7421±0.0016          & 0.7505±0.0087          \\\midrule
    \checkmark             & \checkmark             & \checkmark             & \textbf{VMD}     & \textbf{0.7109±0.0163} & \textbf{0.722±0.0181} & \textbf{0.8191±0.0186} & \textbf{0.7793±0.0077} & \textbf{0.7987±0.0115} \\ \bottomrule
    \end{tabular}
\end{table*}%

\begin{figure}
    \centering
    \includegraphics[width=1\linewidth]{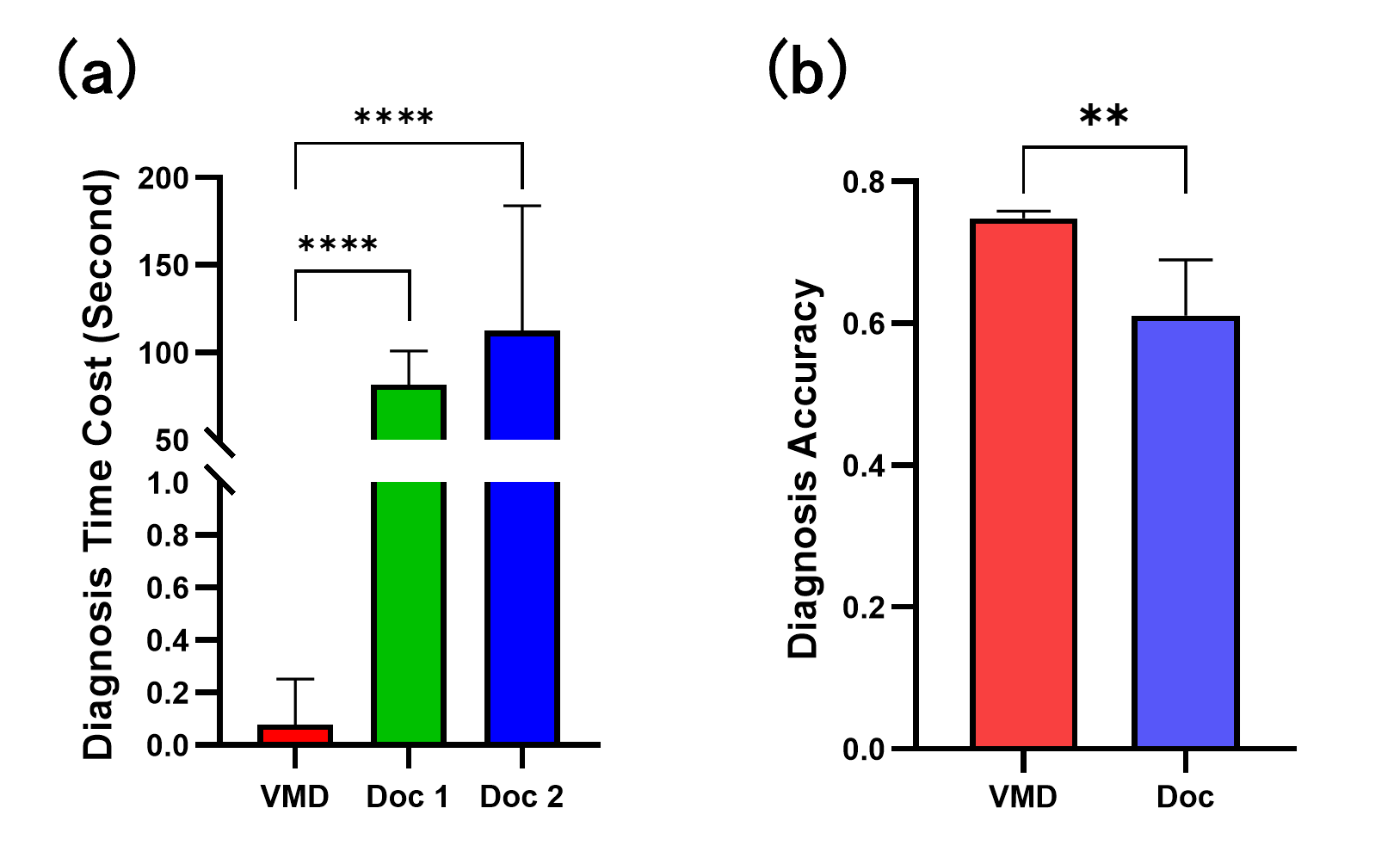}
    \caption{Compare the diagnostic performance of the VMD student model with junior doctors on independent carotid 3D MRI. VMD outperforms radiologists in both time consumption (A) and accuracy (B), with the differences being statistically significant.}
    \label{fig:campare_img}
\end{figure}

The diagnosis process for 27 test images took an average of 2622.5±591.8484 seconds for junior doctors, with each image diagnosis taking 81.6296±19.4188 seconds and 112.6296±71.374 seconds; the average diagnostic accuracy of the two junior doctors was 0.6111±0.0786. Based on the diagnostic results of 5 random seeds, VMD's student model took 20.2628±0.2057 seconds to read the 54 images, 0.7915±0.0826 seconds to load the model, and 17.9684±0.4708 seconds to diagnose, with a total time of 39.0227±0.3771 seconds to finish the diagnosis of 54 cases. The average time to diagnose an image was 0.2659±0.0991 seconds, and the accuracy rate was 0.7481±0.0101, as shown in Fig. \ref{fig:campare_img}.

\subsection{Ablation study} 
In VMD, the multi-level distillation knowledge transfer to the Student comes from both the Teacher and the Expert, and can be divided into two parts based on direct and indirect effects. The direct effects include the feature distribution constraints between the Teacher and the Student, and between the Expert and the Student, which are marked as $\mathcal{I}\left(S,E\right)$ and $\mathcal{I}\left(S,T\right)$ in the methods section, corresponding to (\ref{Eq:elbo 1}) and (\ref{Eq:MIst}). The indirect effects involve the feature distribution constraints through the Expert’s limitation on the Teacher, which is denoted as $\mathcal{I}\left(T,E\right)$ from (\ref{Eq: teacher}). The Teacher learns richer prior knowledge from the Expert, enhancing its diagnostic efficiency and indirectly improving the knowledge transfer level to the downstream Student, achieving stricter feature distribution migration.

To validate the roles of the teacher network and expert network in VMD, we designed two ablation experiments for the teacher network and the student network respectively.

Firstly, in Table. \ref{table:ablation2}, we aim to verify that the knowledge transfer from the Expert can effectively improve the Teacher model’s performance in diagnosing plaque vulnerability in MRI images with limited annotations. We ablated (\ref{Eq: teacher}), and as shown in Table. \ref{table:ablation2}, after the expert network from radiology reports participated in the knowledge transfer during the training phase, the Teacher achieved impressive performance improvements under the dual optimization of variational inference distribution constraints and similarity-based inter-class mutual information optimization. The Teacher model saw more than a 5\% improvement in accuracy, ROC value, recall, and Fbeta, with nearly a 10\% increase in both recall and accuracy. This result indicates that our distillation strategy effectively realizes our hypothesis that the text information in radiology reports, often overlooked in MRI diagnostic intelligent networks, can provide significant assistance to multimodal intelligent diagnostic networks.

Subsequently, we conducted ablation experiments on the diagnostic performance of the Student on unannotated 3D carotid MRI images in the VMD training strategy to demonstrate the positive role of knowledge transfer from the Teacher and Expert networks, as shown in Table. \ref{table:ablation3}. Equation (\ref{Eq:elbo 1}) represents the Expert’s individual guidance to the Student; (\ref{Eq:MIst}) represents the Teacher’s individual guidance to the Student.

On one hand, compared to the baseline method without guidance, i.e., the 3D convolutional diagnostic network simply fine-tuned, the ablation results of both individual guidances showed significant improvements and outperformed the baseline method in all five metrics. This indicates that both guidances provided effective knowledge transfer for improving diagnostic efficiency in the unannotated diagnostic task.

On the other hand, the ablation results of both individual guidances were inferior to the fully guided training strategy of VMD, with VMD achieving more than a 3\% performance improvement in all five metrics. This suggests that the feature transfer from the Teacher and Expert did not result in feature conflicts in improving diagnostic performance, proving the correctness of the inference that both optimization approaches mentioned in the methods section can be summarized as mutual information optimization strategies.

\section{Discussion}
In this study, we propose a novel multi-level cascaded multimodal knowledge distillation network VMD, which is based on MI maximization, providing an intelligent method for the diagnosis of plaque vulnerability in carotid 3D MRI, which relies less on manual annotations by radiologists.
This structure allows for better extraction of feature information beneficial for disease diagnosis from both text and image modalities without introducing additional inference costs. 
Thanks to the diagnostic knowledge derived from cross-modal variational inference and the anatomical prior gained by maximizing inter-class MI, our method achieves the best in most evaluation metrics and is only slightly different from the best results in Recall and Fbeta.
Also, while ensuring diagnostic accuracy, it also has superior diagnostic efficiency.
It can assist junior doctors in the early diagnosis of plaques in patients with carotid stenosis in a real clinical environment. 

In recent years, some research has also provided solutions to plaque detection and component recognition-related problems based on deep learning methods\cite{chen2021domain,WANG2024yufan,wu2019deep}. 
For example, Chen et al.\cite{chen2021domain} constructed a domain adaptive quality assessment and lesion classification model for 3D-MERGE sequence images, but limited the classification target of plaques to the more vague early lesions and late lesions, rather than the clinically more concerned diagnosis of plaque vulnerability; Wang et al.\cite{WANG2024yufan} constructed an end-to-end plaque component segmentation model that can be used in the joint diagnosis process of multiple sequence MRIs, achieving the segmentation and recognition of IPH, LRC, and CA in carotid plaques, but in the inference phase, it requires strict joint images of multiple sequences as inputs. Compared to the above research, VMD focuses more on the full use of more easily obtained multimodal information from reports and images, combined with advanced variational inference optimization ideas, redefines the problem of plaque vulnerability diagnosis as an optimization problem of feature distance, thus using the rich domain knowledge contained in clinically very easy to obtain radiology reports to improve the student model’s understanding of plaque 3D MRI, and at the same time does not need any additional annotations in the inference phase, only need to input the original 3D MRI into the model to get the diagnosis result, which may be easier to deploy in a clinical environment. 

The essence of distillation learning can be understood as the learning of feature distribution between models\cite{KD,AMID,CCL}. 
Some previous student-teacher-based distillation networks focus on learning the mapping across modalities\cite{AMID,CCL,CLIP,tiu2022expert}. 
VMD takes into account the stability of anatomical structures in MRI images and the precise mapping of vascular trends. 
The same type of plaques in the same part have more similar image representations. 
It uses limited annotations of the vessel wall and lumen to provide more local auxiliary information, and combines the similarity optimization method of contrastive learning to narrow the distance of corresponding features in the network of the same type of plaque images, thus achieving further mining of the potential ability of the encoder to process image features. 
In the clinical comparison results in the real environment, it can also be seen that VMD is far superior to junior doctors in diagnostic efficiency, and achieves a higher diagnostic accuracy rate (15\%) compared to junior doctors. 
In the early screening stage, it can give early warning of the patient’s condition, thus assisting the clinical decision-making process for the choice of subsequent treatment methods. 

Our research also has some limitations. 1: Our research included patient data from a single center and the scanning modality was also limited to single-sequence images. In the future, more in-depth research needs to be conducted in multi-center data to eliminate the error impact brought by scanning modalities, parameters, and population. 2: At present, VMD only directly classifies images into vulnerable and stable categories. In the future, the diagnostic target needs to be further expanded, such as diagnosing the risk coefficient by combining more levels of carotid feature information. 3: VMD still includes manual annotation of limited vessel areas during the training phase. In the future, a lower-cost diagnostic model should be constructed by combining multi-sequence complementary information that is beneficial for plaque component identification. 4: The patient data we included were all patients with carotid stenosis. In the future, the proportion of healthy subjects should be expanded to further extend the boundaries of the VMD's ability to automate disease diagnosis.
\section{Conclusions}\label{sec5}

The task of automated diagnosis of vulnerable plaques in carotid 3D MRI is challenging. In this study, we propose a student-teacher-expert distillation strategy named VMD for the automated diagnosis network of carotid 3D MRI, which achieves a significant improvement in diagnosing the vulnerability of atherosclerotic plaques. By combining variational inference and contrastive learning-based MI optimization theory, we efficiently transfer the cross-modal information in the teacher network and the expert network to the student network, without introducing additional inference costs.


\bibliographystyle{IEEEtran}
\bibliography{tmi_vmd}
\appendices



\end{document}